\documentclass[11pt,letterpaper]{article}
\usepackage{main}
\usepackage{times}
\usepackage{latexsym}
\usepackage[pdftex]{graphicx}
\usepackage{color}
\title{Supervised Hierarchical Classification for Student Answer Scoring}

\author{Itziar Aldabe \hspace{6px} Oier Lopez de Lacalle \hspace{6px} I\~{n}igo Lopez-Gazpio\footnotemark \hspace{6px} Montse Maritxalar\\	
	IXA NLP group\\
	University of the Basque Country\\
	Manuel Lardizabal 1, 20018 Donostia\\
	{\tt \{itziar.aldabe\},\{oier.lopezdelacalle\},\{inigo.lopez\},\{montse.maritxalar\}@ehu.eus}
	}

\date{}

\begin{document}

\maketitle
\footnotetext{Contact for paper related queries.}
\begin{abstract}

This paper describes a hierarchical system that predicts one label at a time for automated student response analysis.
For the task, we build a classification binary tree that delays more easily confused labels to later stages using hierarchical processes.
In particular, the paper describes how the hierarchical classifier has been built and how the classification task has been broken down into binary subtasks.
It finally discusses the motivations and fundamentals of such an approach.

\end{abstract}

%%% Local Variables: 
%%% mode: latex
%%% TeX-parse-self: t
%%% TeX-PDF-mode: t
%%% TeX-master: "main"
%%% End: 

\section{Introduction}
\label{sec:introduction}

One of the aims of educational natural language processing is to provide useful feedback to students in Learning Management Systems.
In this sense, \newcite{dzikovska-EtAl:2013:SemEval-2013} proposed the Semeval-2013 Student Response Analysis (\textit{SRA}) task to automatically grade open question answers.
The SRA task introduced different levels of granularity, and the results of different systems showed that the higher the number of categories, the worst the overall results were.
For example, regarding the best results, the macro-average F1 value dropped from 0.72 in the 2-way task (2 categories) to 0.42 in the 5-way task (5 categories). 

\begin{figure}[!hbt]
 \begin{center}
   \includegraphics[scale=0.4]{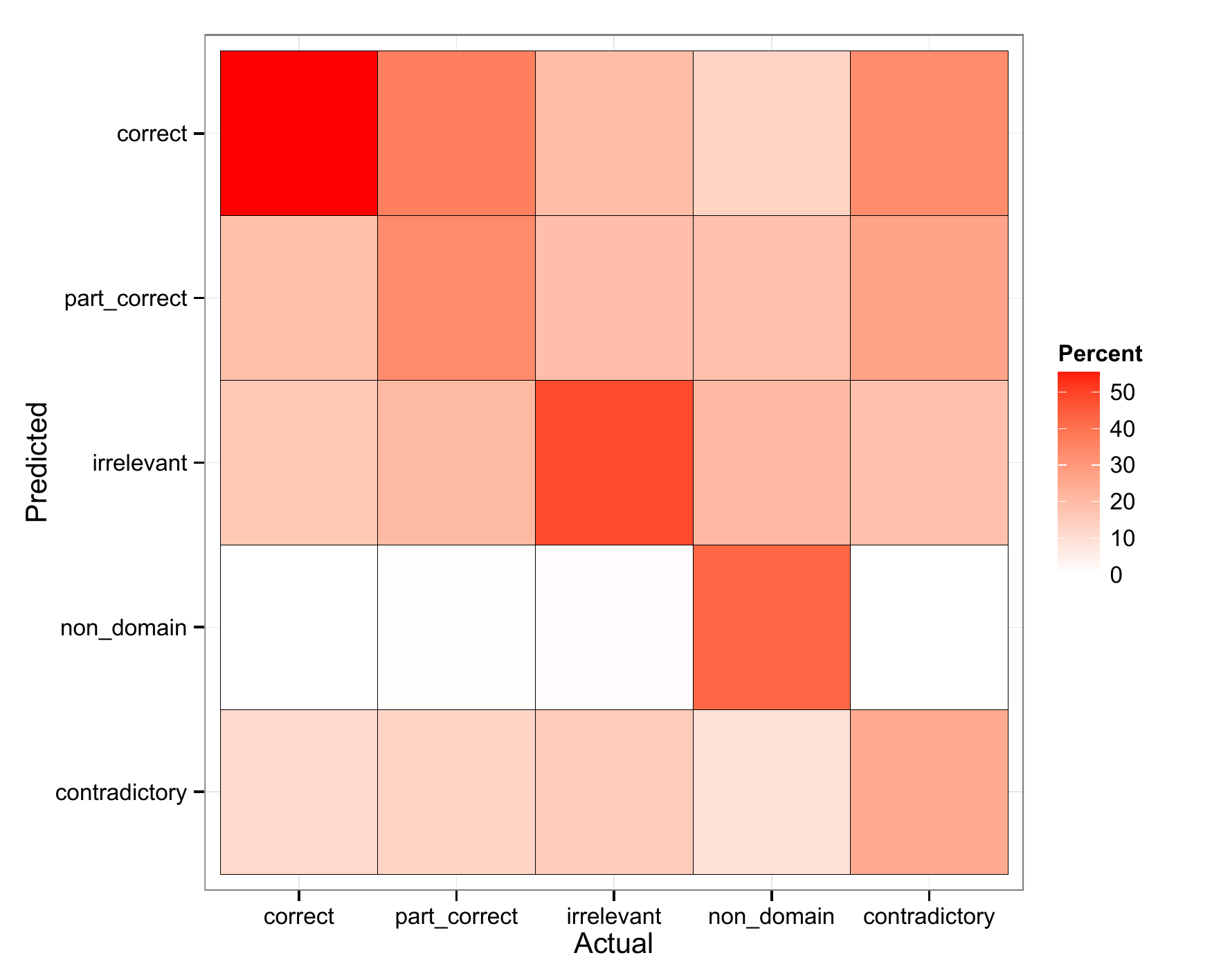}
 \end{center}
  \caption{Confusion matrix of a 5-way supervised classifier on the Semeval-2013 training data.}
  \label{fig:cm}
\end{figure}

Our hypothesis states that the set of categories defined for the task in \newcite{semeval_previous_work} 
presents complex and gradual relationships.
As a consequence, these relations may be indicative of potential weakness to score the answers automatically.
Given a set of labels (correct, partially\_correct, irrelevant, non\_domain, contradictory) intuition dictates that we can expect major problems
differentiating {\tt correct} and {\tt partially\_correct} answers, than {\tt correct} and {\tt non\_domain}\footnote{{\tt non\_domain}  answers do not include question's domain content.} answers.
Within this context, the confusion matrix of a tuned 5-way classifier (see figure \ref{fig:cm}) gives us an appropriate way to automatically identify the problematic label-relationship in order to set fruitful steps towards a better solution.
Actually, the confusion matrix shows higher error rate when discarding between {\tt correct} and {\tt partially\_correct} answers, than between {\tt correct} and {\tt  non\_domain} answers.

\begin{figure}[!hbt]
 \begin{center}
   \includegraphics[scale=0.4]{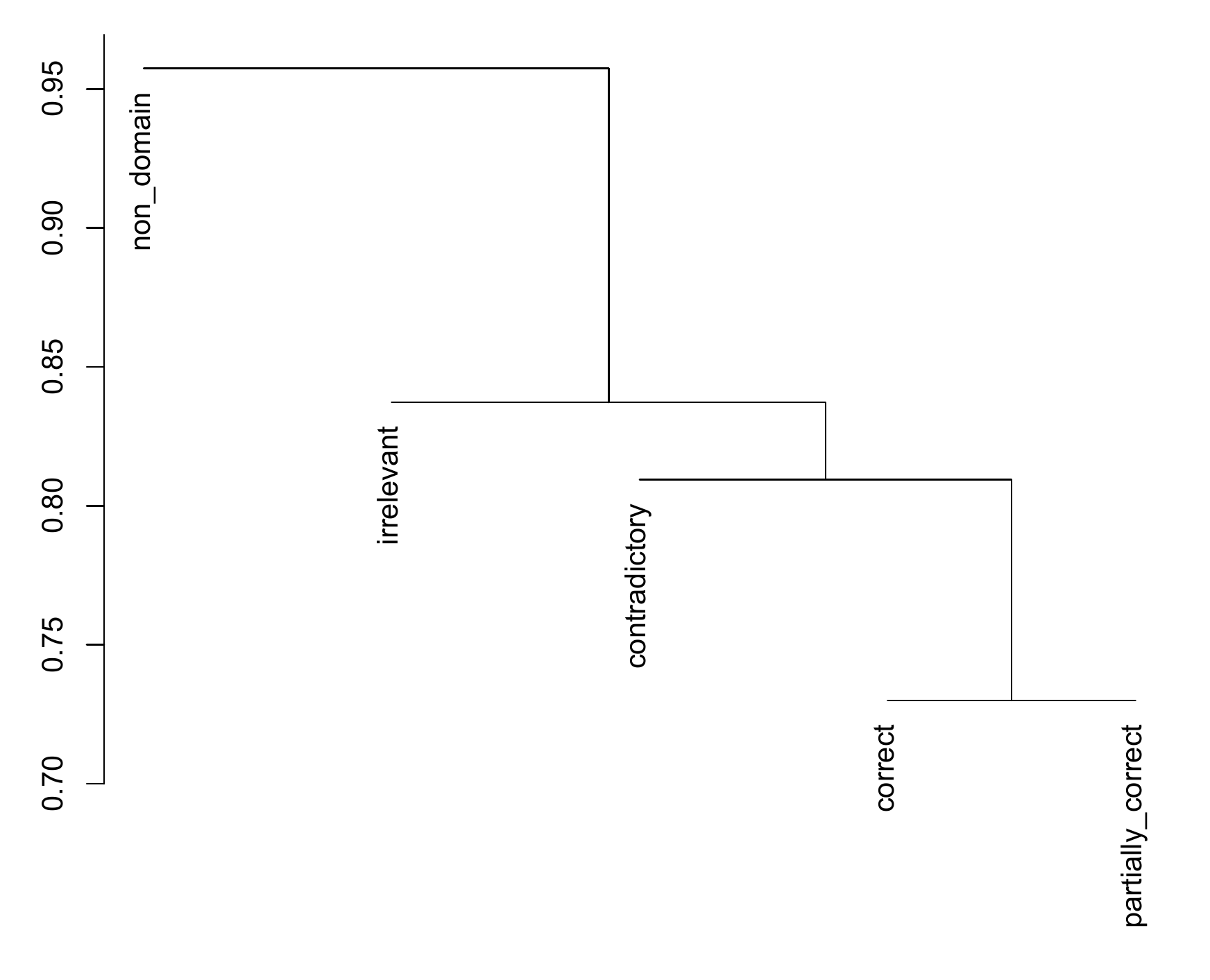}
 \end{center}
 \caption{\label{agglomerative_hierarchical_clustering} Agglomerative hierarchical clustering among the Semeval-2013 data. }
\end{figure}

Previous work has shown that binarizing multiclass problems can be beneficial for improving the overall performance of some systems \cite{lei2005half} \cite{Allwein:2001:RMB:944733.944737,lei2005half,marszalek2008constructing}.
In this work, we present a method which walks through distinct nodes by making binary decisions at each step.
Thus, decomposing the main task in several subtasks.
SRA systems participating on Semeval-2013 faced the problem as a 5-way classification task.
They assumed no hierarchy among the labels and ignored the difficulties which rely on the structured nature of the category set.

On the contrary, we construct label-partitions and sort labels according to the confusion-relatedness (see figure \ref{agglomerative_hierarchical_clustering}).
To the best of our knowledge, this is the first SRA system that tries to take advantage from the analysis of the label interdependencies.
In sum, our work tries to induce structured information so that 1) we can prioritize over the label relationships and 2) we can decompose the main task in simpler binary subtasks.
\section{Problem formulation and data description}
\label{sec:probl-form-flat}

The Semeval-2013 SRA task addresses the problem of grading student responses from different science domains.
The corpora for this task has been created out of two established sources:
the BEETLE corpus, a data set collected and annotated during the evaluation of the BEETLE II tutorial dialogue system \cite{dzikovska2010beetle};
and the SCIENTSBANK corpus, a set of student answers to questions from sixteen science modules in the Assessing Science Knowledge Assessment Inventory \cite{nielsen2008annotating}.
The objective of the task is to determine the label of the answer given an open question and a reference answer.
Student answers can be classified at different levels of granularity: 5-way task, 3-way task and 2-way task.
In the 5-way task, the one we follow in this work, the aim is to classify the correctness of the answers as correct, partially correct or incomplete, contradictory, irrelevant, and not in the domain.
The whole task is completely documented in \newcite{dzikovska-EtAl:2013:SemEval-2013}.

To build, train and test our hierarchical system we use the same training and test sets as the \textit{EHUALM} system does.
EHUALM is an ensemble of distinct supervised classifiers that took part in the Semeval-2013 5-way task.
It ranked considerably well above mean and median in different scenarios of the task.
The full system is described in \newcite{ehualm}.
As regards the dataset, which is included in the supporting material, 
it comprises a total of 30 syntactic and semantic similarity features computed between the question and the reference answer(s); but also, between the question and the student answer.
The syntactic and semantic similarity features can be grouped into the following sets: text overlap features, WordNet-based features, graph-based features, corpus-based features and predicate-argument features.

%%% Local Variables: 
%%% mode: latex
%%% TeX-parse-self: t
%%% TeX-PDF-mode: t
%%% TeX-master: "main"
%%% End: 
\section{Hierarchy based system}

% main idea and purpose
In section \ref{sec:introduction} we stated that the inner relation between labels is meaningful to model similarities among SRA.
Such a relation can be used to build a hierarchical approach in order to consider the label relationships inside the classifying model.
We consider two labels to be similar if the output of a tuned 5-way classifier confuses them frequently.
In other words, we expect higher difficulties among labels that are semantically close.

In order to obtain relatedness values among the labels of the Semeval task we need to construct a distance matrix.
The distance matrix will then be used to construct a dendrogram that connects distinct labels according to their relatedness (see figure \ref{agglomerative_hierarchical_clustering}).
As a first step we have calculated the similarity between two labels as follows:
given two labels ($i$ and $j$) we count how many times one label ($i$) is assigned when the other is the correct one ($j$), and viceversa.
The average of the two normalized counts defines the similarity between both labels.
Similarity is transformed into distance as $dist(i,j) = 1 - sim(i,j)$.
Note that the matrix is square and symmetric, and each element at position ($i$, $j$) determines the distance between labels $i$ and $j$.
Once obtained the distance matrix, we apply agglomerative hierarchical clustering to obtain the partitioning.
The code and data to produce the hierarchical structure is supplied in the supporting material.

Once the hierarchical structure is defined, a binary multi-class Support Vector Machine hierarchy \cite{Chang:2011:LLS:1961189.1961199} hierarchy
is built (2-way SVM tree) by recursively dividing the training dataset of K classes into two subsets of classes.
Thus, each node decides if a sample belongs to a specific label or to one of the sub-hierarchy labels.

For the experiments we define two different hierarchical configurations: H1 and H2.
\textbf{H1} starts discarding the most disimilar labels (i.e. {\tt non domain} \emph{vs} {\tt rest}) and finishes with the most similar ones (i.e. {\tt correct} \emph{vs} {\tt partially correct}).
In contrast, \textbf{H2} starts from the most similar labels and walks into the most disimilar ones.

Each node of the tree is trained independently by mapping the whole training set to the classes it needs to handle.
The leaf label is the positive class (e.g. {\tt non domain} on the top of figure \ref{agglomerative_hierarchical_clustering}) and the remaining labels under the sub-hierarchy make up the negative class.
This approach is similar to the {\it all-pairs} approach mentioned in \newcite{Allwein:2001:RMB:944733.944737} as some instances are ignored when training the non-top binary classifiers.
That is, when training the non-top binary classifiers we do not consider the instances of the classes handled in higher levels of the hierarchy.
For testing we apply the whole hierarchy to each incoming instance.
We evaluate the architecture in two ways: 1) measuring the local performance of each level, and 2) measuring the overall performance of the tree.

%%% Local Variables: 
%%% mode: latex
%%% TeX-parse-self: t
%%% TeX-PDF-mode: t
%%% TeX-master: "main"
%%% End: 
\begin{table}[!hbt]
\begin{center}
\scriptsize
\begin{tabular}{|l|l|c|}
  \hline
  {\bf System}	&{\bf Description}		&{\bf F-Score} \\
  \hline
  H1 overall	&Whole tree		&0.56   \\
  H1 L1		&Non domain vs Rest	&0.988  \\
  H1 L2		&Irrelevant vs Rest	&0.856  \\
  H1 L3		&Contradictory vs Rest	&0.795  \\
  H1 L4		&Partially vs Correct	&0.745  \\
  \hline
  {\bf System}		&{\bf Description}		&{\bf F-Score} \\
  \hline
  H2 overall	&Whole tree		&0.568		\\
  H2 L1		&Correct vs Rest	&0.762		\\
  H2 L2		&Partially vs Rest	&0.699		\\
  H2 L3		&Contradictory vs Rest	&0.78		\\
  H2 L4		&Irrelevant vs Non domain&0.96		\\
  \hline
  
\end{tabular}
\end{center}
\caption{\label{hierarchy_results}  Micro F-Scores on 5 fold cross-validation for each level ('L') of the two hierarchical configurations ('H').}
\end{table}

\section{Experiments and results}
To train and evaluate the two hierarchical configurations we have used the SRA train and test data described in section \ref{sec:probl-form-flat}.
Each binary SVM classifier of the hierarchy has been tuned using 5-fold cross validation, maximizing the micro accuracy as the objective function.
We have calculated level-wise and overall results in order to analyze both:
the contribution of each binary classifier, and the overall results.
Table \ref{hierarchy_results} summarizes the results obtained for both hierarchical configurations.

Concerning the level evaluation, the highest F-score value (0.988) is obtained when dealing with instances of the most distanced class (top level of the H1 configuration).
Accuracy decreases as descending on the hierarchy.
Just the opposite effect is shown in the other hierarchy, as it deals first with the most similar classes.
Actually, H2 obtains the highest F-score value (0.96) in the lowest level.
This result has much to do with the confusion matrix shown in figure \ref{fig:cm} as it is the class with lowest false positive rate ({\tt non\_domain} predicted row). 
As regards overall results, accuracy drops considerably in comparison to the values obtained in the level results.
In our opinion, this result is related to the \emph{error propagation} in the tree.
We define the error propagation as the instances that really belonging to one class are incorrectly classified and continue down the tree to the next level, degrading the performance in subsequent levels.

Comparing performance among the hierarchical approaches and a 5-way flat SVM model (also tuned on the training set) results turn to be similar.
The micro F-score results for H1 and H2 are 0.56 and 0.568 respectively (0.571 for the flat SVM),
and the macro F-score results for H1 and H2 are 0.553 and 0.566 respectively (0.566 for the flat SVM).

Finally, we evaluated the H2 system\footnote {We present the results of H2 which outperformed H1.} using the test data provided for the SRA task.
The micro F-score obtained for H2 in the test set was 0.426 (0.415 for the flat SVM) and the macro F-score was 0.408 (0.39 for the flat SVM).
Contrary to what happened in the cross-validation setting, the hierarchy based approach outperformed the SVM model.
This indicates that the hierarchy based approach prevents from overfitting as the hierarchy structure imposes coherent biases that correlates the problem formulation and label relationship.

%%% Local Variables: 
%%% mode: latex
%%% TeX-parse-self: t
%%% TeX-PDF-mode: t
%%% TeX-master: "main"
%%% End: 

\section{Discussion}

The present paper describes ongoing work on building hierarchy based models to automatically grade student answers as an alternative to typical N-way models.
We show that the elucidation of the relationship between classes in the Semeval-2013 task is more difficult than it could be expected using basic hierarchical structures.
Even promising results are obtained at level-wise training, the overall accuracy is deteriorated when the whole system is combined.
Nevertheless, results show that the hierarchical approach outperforms the flat 5-way SVM at test.

In order to address this issue, we consider that a deeper label interdependency analysis is necessary to effectively discriminate among classes with high confusion rates.
That is why as a first step we conducted an error analysis and found crucial for future work to specialize the feature set being used for training.
Actually, we think that specific features may considerably contribute to certain levels of the hierarchy.
For instance, features that explicitly handle negation are critical when classifying contradictory instances, but useless in other stages.

As regards the error analysis we have taken the confusion matrix of the whole tree performance and analyzed the incorrectly classified major groups.
Out of the total error rate of H2 (42\% of instances incorrectly classified) the most problematic misclassifications are distributed across the following categories, which we briefly describe:
a) {\tt partially\_correct} instances being classified as {\tt correct}: 18\% error rate;			 %658 / 3761 
b) {\tt correct} instances being classified as {\tt partially\_correct}: 14\% error rate; 			 %515 / 3761
c) {\tt contradictory} instances being classified as {\tt correct}: 11\% error rate;  			 %412 / 3761
d) {\tt contradictory} instances being classified as {\tt partially\_correct}: 9\% error rate; 		 %341 / 3761 
e) {\tt irrelevant} instances being classified as {\tt correct}: 8\% error rate.  				 %285 / 3761

The mentioned five cases account for the 60\% of the total error rate, while all of the other error cases account each for less than 6\% error.
Our analysis concluded that, as expected, most of the errors require not only general lexical features but also more fine grained ones so that to be correctly classified.
For instance, we think that the approach taken in the interpretable pilot described in \newcite{agirre2015semeval} can be effective for cases
a), b) and e) so that to obtain a more fine grained linkage between the concepts of the student answer and the concepts of the reference answer.
The usefulness of this approach resides in making alignments between concepts in the reference-student pair.

On the contrary, for {\tt contradictory} related misclassifications, such as: c) and d) we think that specific features able to handle negation are crucial to improve performance.
In fact, we perform a deeper analysis of the c) case by randomly samplig 20 instances and found that even for humans the annotated gold values are ambiguous certain times.
Out of the 20 student answers analyzed we found 5 ambiguous cases, some of them related to the usage of negative polarity particles.
For example, for the question \textit{"Why was bulb C off when switch Z was open?"} and reference answer \textit{"There was no longer a closed path containing Bulb C and the battery"}
the following student answer is annotated as { \tt contradictory}: \textit{"switch Z created a gap in the closed circuit required for bulb C"}.
Though, our hierarchical system scores it as {\tt correct}, which we think is the most suitable grade.
Just the same happens with the following reference-student pair: \textit{"(Reference) The more blocks the truck carries, the less distance the truck travels in 10 seconds."}
and \textit{" (Student) It went farther with less blocks and it went no farther with more blocks."}

In all, to grade student answers is a challenging task that requires to analyze and identify errors and misconceptions based on reference responses.
As regards student answer clustering, even some work has been done, such as the work described in \newcite{basu2013powergrading} and \newcite{zesch2015reducing} 
it is still an open research line to prove whether clustering structures can meaningfully improve performance.

%%% Local Variables: 
%%% mode: latex
%%% TeX-parse-self: t
%%% TeX-PDF-mode: t
%%% TeX-master: "main"
%%% End: 
\section{Future work}

Even if the hierarchical decomposition explained has showed negative results, we plan to continue analyzing distinct configurations to specialize the concrete feature list for each level of the hierarchy.
Actually, we think that the system requires new attributes in order to improve.
Distinct attributes may contribute differently as regards each level, for instance, features that explicitly handle negation seem to be critical in the modules responsible for classifying contradictory instances, but useless in other modules.
Moreover, we also plan to explore new similarity features of top performing systems of the SemEval task that could contribute to our system.
In all, we plan to continue analyzing strategies and developing techniques in order to improve the provided automatic scores.

%%% Local Variables: 
%%% mode: latex
%%% TeX-parse-self: t
%%% TeX-PDF-mode: t
%%% TeX-master: "main"
%%% End: 
\section{Acknowledgments}

This project was partially funded by the national program for university professor and philosophy doctor training (FPU13/00501).

%%% Local Variables: 
%%% mode: latex
%%% TeX-parse-self: t
%%% TeX-PDF-mode: t
%%% TeX-master: "main"
%%% End: 

\bibliographystyle{main}
\bibliography{main}

\end{document}